\documentclass[letterpaper, 10pt, conference]{ieeeconf}   % Comment this line out if you need a4paper

\IEEEoverridecommandlockouts                              % This command is only needed if 
\let\labelindent\relax                                    % Solve name clash for enumitem package

\usepackage{graphicx}               % For pdf/bitmapped graphics files
\usepackage{amsmath}                % Amsmath package for maths and equations
\usepackage{amssymb}                % Maths symbols
\usepackage{amsfonts}               % To get the \mathbb alphabet
\usepackage{enumitem}               % For customisation of enumerate environments
\usepackage{siunitx}                % For SI units
\usepackage{url}                    % For URL typesetting and line breaking
\usepackage{xspace}                 % For smart spaces in commands
\usepackage[T1]{fontenc}            % For special characters such as '_'
\usepackage[hidelinks]{hyperref}    % For hyperlinks (citation hyperlinks won't work with ieeeconf)
\usepackage{textcomp}               % For \textregistered
\usepackage{algorithm, algorithmic} % For pseudocode boxes
\usepackage{array}                  % For pseudocode boxes
\usepackage{eqparbox}               % For pseudocode boxes
\makeatletter
\let\MYcaption\@makecaption
\makeatother
\usepackage[font=footnotesize]{subcaption}
\makeatletter
\let\@makecaption\MYcaption
\makeatother

\graphicspath{{images/}}
\DeclareGraphicsExtensions{.pdf,.jpeg,.png,.jpg}

\algsetup{linenosize=\small}

\newcommand{\degreem}{^{\circ}} % Degree circle

\newcommand{\seclabel}[1]{\label{sec:#1}}

\newcommand{\figlabel}[1]{\label{fig:#1}}

\newcommand{\tablabel}[1]{\label{tab:#1}}

\newcommand{\secref}[1]{(Sec.~\ref{sec:#1}\xspace)}

\newcommand{\figref}[1]{Fig.~\ref{fig:#1}\xspace}
\newcommand{\tabref}[1]{Table~\ref{tab:#1}\xspace}

\newcommand{\iguhop}{igus\textsuperscript{\tiny\circledR}$\!$ Humanoid Open Platform\xspace}

\newcommand{\degree}{$\degreem$\xspace}

\title{\LARGE \bf Online Visual Robot Tracking and Identification \\ using Deep LSTM Networks}

\author{Hafez Farazi and Sven Behnke% <-- This % stops a space
\thanks{All authors are with the Autonomous Intelligent Systems (AIS) Group, Computer Science Institute VI,
        University of Bonn, Germany. Email: {\tt\small farazi@ais.uni-bonn.de}. This work was partially funded
        by grant BE 2556/10 of the German Research Foundation (DFG).
        }}

\usepackage{eso-pic}

\AtBeginDocument{\AddToShipoutPictureFG*{\AtTextUpperLeft{\put(0,\LenToUnit{40pt}){\parbox{\textwidth}{\centering\bfseries
IEEE/RSJ International Conference on Intelligent Robots and Systems (IROS), Vancouver, Canada, 2017
}}}}}

\begin{document}

% Typeset the paper title
\maketitle
\thispagestyle{empty}
\pagestyle{empty}

%%%%%%%%%%%%%%%%%%%%%%%%%%%%%%%%%%%%%%%%%%%%%%%%%%%%%%%%%%%%%%%%%%%%%%%%%%%%%%%%
\begin{abstract}

Collaborative robots working on a common task are necessary for many 
applications. One of the challenges for achieving collaboration in a team of 
robots is mutual tracking and identification. We present a novel pipeline for 
online vision-based detection, tracking and identification of robots with a 
known and identical appearance. Our method runs in real-time on the limited 
hardware of the observer robot. Unlike previous works addressing robot tracking 
and identification, we use a data-driven approach based on recurrent neural 
networks to learn relations between sequential inputs and outputs. We 
formulate the data association problem as multiple classification problems. A deep LSTM 
network was trained on a simulated dataset and fine-tuned on small set of real 
data. Experiments on two challenging datasets, 
one synthetic and one real, which include long-term occlusions, show promising 
results.

\end{abstract}
%%%%%%%%%%%%%%%%%%%%%%%%%%%%%%%%%%%%%%%%%%%%%%%%%%%%%%%%%%%%%%%%%%%%%%%%%%%%%%%%
\section{Introduction}
Multi-target tracking is a challenging and well-known problem in computer 
vision, which has been studied for decades~\cite{zhang2012robust,yang2005fast,allen2004object}. In multi-target 
tracking, we find objects of interests, assign them a unique ID, and follow them 
over time. Multi-target tracking is used in many applications including 
automated surveillance and traffic monitoring. Tracking by detection is one of 
the most common approaches, which uses a detector to discard unnecessary 
information from the video sequence, and reduces the problem to data association 
for a smaller discrete set of detections. The data association problem, 
especially in cluttered environments and with multiple closely spaced and 
possibly occluded objects, is one of the main reasons that multi-target tracking 
is a fundamentally harder problem than single-target tracking. Furthermore, the 
number of visible targets may be unknown and vary over time. Initiation and 
termination of the tracks should be robust to false positives and false 
negatives. Due to the aforementioned difficulties, state of the art results are 
still far from human-level accuracy~\cite{leal2015motchallenge}.

In this work, we address tracking and identification of multiple robots of identical appearance,
which is a problem with an additional level of difficulty. 
Moreover, we are not using the internal location estimate calculated by each robot, so 
that the system is usable even in situations when robots are not localized. Despite the lack of visual cues, our system is able to track the target robots, and in addition identify which detection corresponds to which exact robot. This 
is done using a deep Long Short-Term Memory (LSTM) network, based on a set of 
detections that include heading estimates from visual observations and heading information provided by 
the robots. In our application, the output of the system, which is the 
estimation of the relative location and heading of each observed robot, is broadcasted to the observed robots
for further use in improving 
self-localization or high-level cooperative behavior. The challenging nature of 
our setup suggests that the proposed method is also suitable for supporting 
other robot collaboration tasks.~\figref{overview} gives an overview of our 
system. A video of the experiment is available at our 
website\footnote{\url{http://www.ais.uni-bonn.de/videos/IROS_2017_LSTM}}.

Although in many application areas,--- from computer vision to machine 
translation,--- deep learning approaches are shaping the state-of-the-art, in 
multi-target tracking and data association problems, there are surprisingly few works. As identified by Milan 
et al.~\cite{Milan:2017:AAAI_RNNTracking}, 
the two most notable reasons for this are the lack of available training data,
and the large amount of generalization required by the network to account for the variability in the data.
This includes the variability in the viewpoints and length of sequences, and the unknown
cardinality of the input.
%-----------------------------------
\begin{figure}[t]
\vspace*{1ex}
\centering
\hspace*{-0.2cm}
\includegraphics[width=10.9cm]{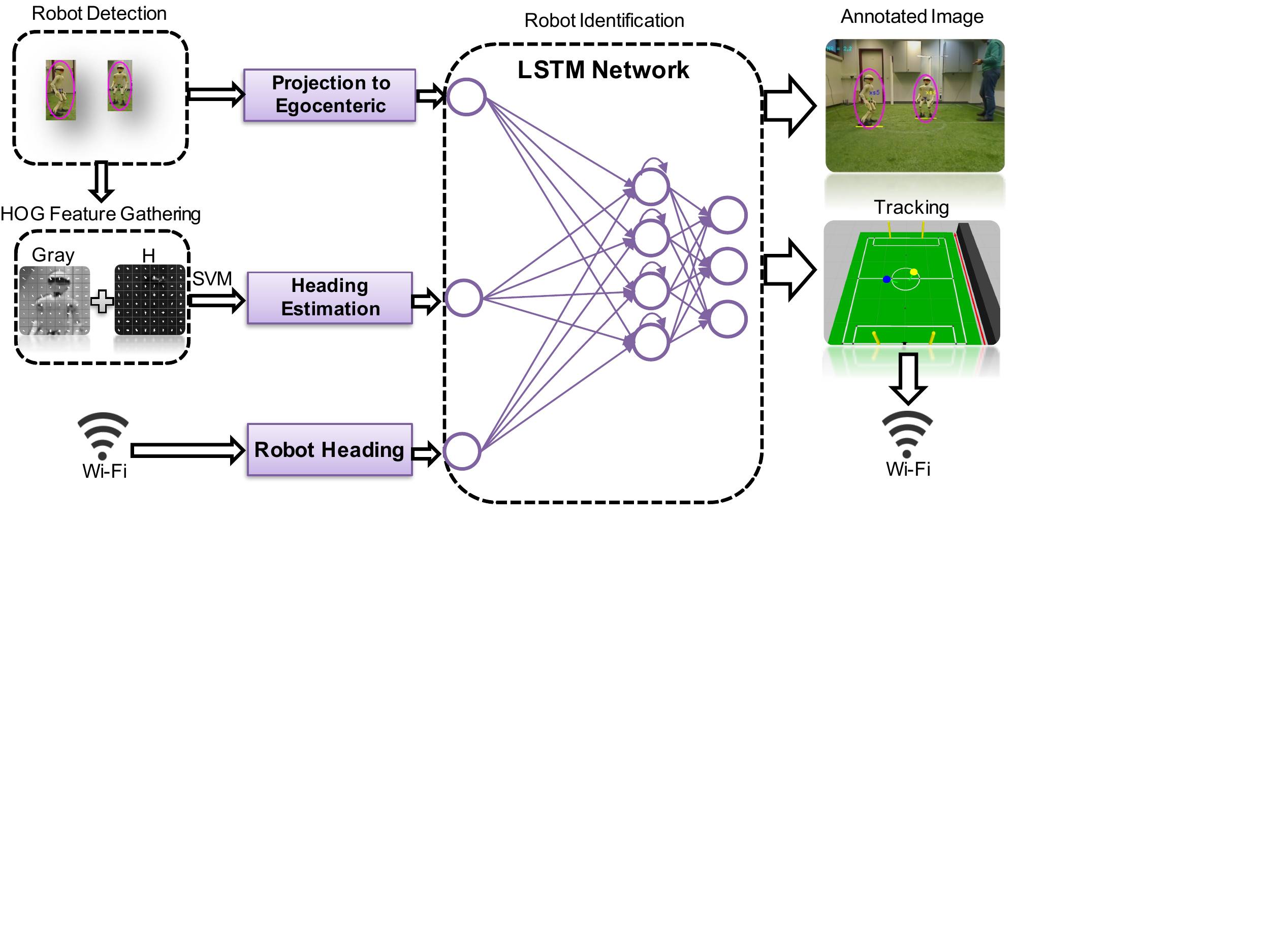}
\vspace*{-27.5ex}
\caption{Overview of our approach. After robot detection, the heading of each robot is estimated based on HOG features. Robot identification is done using LSTM network by matching observed robot tracks to headings reported by the robots using Wi-Fi. The calculated location is then broadcasted by Wi-Fi to the robots for further use.}
\figlabel{overview}
\vspace*{-4.5ex}
\end{figure}
%-----------------------------------
The main contributions of this paper include:
\begin{enumerate}
\item The introduction of a novel pipeline to visually identify, track and localize a set of 
identical robots in real-time,
\item The use of a single RNN for the complete task, including data association, initiation and termination
of targets, without prior knowledge about the environment, robot dynamics and occlusions, and
\item The introduction of a generative model that can sample an arbitrary number of data, allowing
the dynamics of real environments to be learnt largely from a simulated dataset.
\end{enumerate}

%%%%%%%%%%%%%%%%%%%%%%%%%%%%%%%%%%%%%%%%%%%%%%%%%%%%%%%%%%%%%%%%%%%%%%%%%%%%%%%%
\section{Related Work}
We divide the discussion of related works into three categories, \emph{multi-target tracking}, 
\emph{deep learning}, and \emph{robot detection and tracking}.

\textbf{Multi-target tracking} with no category information is referred to as 
category-free tracking (CFT)~\cite{zhang2012robust}. With the use of manual 
initialization, CFT approaches typically do not require a pre-trained detector. By 
discriminating other regions of the images, CFT methods work mainly based 
on visual appearance. Two notable approaches, without deep learning, are works 
by Yang et al.~\cite{yang2005fast} and Allen et al.~\cite{allen2004object}. CFT 
methods are usually computationally inexpensive, but they are prone to drift and 
cannot easily recover from occlusions.

Another popular type of tracking methods is association based tracking (ABT)~\cite{xing2009multi}, which works by means of a discrete set of detections. In contrast 
to CFT, this approach does not suffer from extreme model drifts. In ABT, 
continuous target detections are linked over time to form tracks. In many works, 
the probability of association is calculated based on a fixed motion model or 
visual similarity. Global track association is then computed either with the
Hungarian method~\cite{xing2009multi}, a Conditional Random Field (CRF) 
\cite{milan2015joint}, or a Markov chain Monte Carlo method~\cite{yu2007multiple}.

Joint probability data association (JPDA)~\cite{fortmann1983sonar} was 
originally developed for radar and sonar tracking, and for a long time was 
considered too computationally expensive for computer vision applications. With 
proper approximation~\cite{hamid2015joint} and a novel appearance model 
\cite{kim2015multiple}, JPDA has recently found use in multi-target tracking.

Sousa et al.~\cite{sousa2013human} proposed a non-visual human tracking and 
identification method using a sensing floor and wearable accelerometer. They 
exploited room entrance time and trajectory association for tracking and 
identification. Perez-Escudero et al.~\cite{perez2014idtracker} proposed the 
generation of a visual target fingerprint, to track and identify targets based 
on appearance differences. For multi-person tracking, Maksai et al.~\cite{maksai2016globally} extracted behavioral patterns from the ground truth 
and then used them to guide the tracking algorithm.

\textbf{Deep Learning} approaches have shown successful results in a number of 
application domains--- from speech recognition~\cite{dahl2012context} to visual 
classification~\cite{krizhevsky2012imagenet}. In these approaches, a large 
number of parameters, which are designed to capture the hierarchical 
representation of the data, are automatically tuned based on a large amount 
of data. Two related works by Ondruska et al.~\cite{ondruska2016deep}~\cite{ondruska2016end} use deep learning approaches for tracking. Note that 
they are using 2D laser scanner data in pixel coordinates, making it unsuitable 
for our application. The first of the two works~\cite{ondruska2016deep} was only 
tested on simulated data with a constant velocity motion model. So although 
Ondruska et al. got very promising results in an unsupervised fashion, their 
work is not applicable in our setting and on real noisy data with a more complex
underlying motion model. Another more recent work is a unified RNN structure for
multi-tracking proposed by Milan et al.~\cite{Milan:2017:AAAI_RNNTracking}. 
Although the authors report compelling results, it is not straightforward to adapt their work to our 
problem. They utilized a multi-stage pipeline that needs to be trained separately for different tasks
like predictions, update, birth/death control, and data association. Moreover, the approach works in image
coordinates and it is not clear how performance would change for a different viewpoint. 
To make it work well in different camera positions, one would need to give the network enough samples from various
camera angles to force learning different camera transformations. A CFT-based approach using a
convolutional neural network (CNN) for single target tracking has been proposed by Wang et al.~\cite{wang2017robust}.
One of the best results on the multi-object tracking (MOT) benchmark is a method recently proposed by Sadeghian et al.
\cite{sadeghian2017tracking} that is based on appearances and motion models using CNN and deep RNNs. Their method is
not applicable in our setting mainly because of the appearance model used.

\textbf{Robot Detection and Tracking} targeted for robot soccer has been done by 
Marchant et al.~\cite{marchant2013cooperative} using a combination of ultrasonic 
and visually perceived data. Note that most of the state-of-the-art object 
detection approaches~\cite{redmon2016yolo9000,liu2016ssd} cannot be used 
in our setup, due to the absence of a GPU, and the limitations in computing 
power on the observer robot. Arenas et al.~\cite{arenas2007detection} proposed a 
real-time capable Aibo and humanoid robot detector using a cascade of boosted 
classifiers. In a more recent work, Ruiz-Del-Solar et al.~\cite{ruiz2010visual} 
proposed nested cascades of boosted classifiers for detecting legged robots. In 
the follow-up work from Ruiz-Del-Solar et al.~\cite{ruiz2010play}, the gaze 
direction of a humanoid robot is estimated using a scale invariant feature 
transform (SIFT) descriptor.

The authors of this work have previously used color segmentation for humanoid robot detection 
in the context of RoboCup soccer~\cite{Farazi2015}. In another related work 
pertaining to humanoid robot tracking and identification, the authors exploited 
a Kalman filter for motion modeling, and the Hungarian method for tracking and 
identification~\cite{farazireal}.

%%%%%%%%%%%%%%%%%%%%%%%%%%%%%%%%%%%%%%%%%%%%%%%%%%%%%%%%%%%%%%%%%%%%%%%%%%%%%%%%
\section{Problem Formulation}

A robot, or possibly a stationary camera as a substitute, is used to observe a collection
of $N$ moving robots with the same appearance. Ideally, each visible 
robot should be tracked and identified by the observer in each frame. In practice,
for arbitrarily short or long durations, each of the robots can be fully or partially visible;
or not visible at all. The observed robots can perform a multitude of possible actions, including
but not limited to walking, kicking, standing, and getting up. With the use 
of an internal 9-axis inertial measurement unit (IMU), each robot calculates
and broadcasts their absolute heading direction over Wi-Fi. Wi-Fi quality is not ensured
 and can have delays or even data loss. We utilized the NimbRo Network library~\cite{nimbro}
for increasing the robustness of Wi-Fi communications. The system is designed to detect, track and
identify $N$ moving robots, solely based on images captured by the observer, and the received heading information.

%%%%%%%%%%%%%%%%%%%%%%%%%%%%%%%%%%%%%%%%%%%%%%%%%%%%%%%%%%%%%%%%%%%%%%%%%%%%%%%%
\section{Vision System}
%TODO: VIDEO link
\subsection{Robot Detection}
Unlike the existence of pre-trained detectors for pedestrians or animals, there is 
no robot detector that can work out of the box for our application. Hence, we 
have designed and implemented an \iguhop robot detector that can work robustly 
under various robot configurations and lighting conditions. We expect that our 
proposed detector can work for other robots with proper retraining.

We used Histograms of Oriented Gradients (HOG)~\cite{zhu2006fast} features
because they are computationally efficient and invariant to changes of
illumination. In contrast to the popular pedestrian detection~\cite{dalal2006object},
which uses support vector machine (SVM) with sliding-window, we saved computational cost
and used a cascade of rejectors with the AdaBoost technique to choose which features to
evaluate in each stage. Our detector is similar to Zhu et al.~\cite{zhu2006fast}. HOG features
are not rotation and scale invariant, so we apply random transformations with normal distributions
to expand the number of images collected by the user.

We restrict random rotations to \textpm\SI{15}{\degree}, to give the 
classifier the chance to learn the shadow under the robot. We also emulated 
partial occlusions by randomly cutting some portion of the positive samples. For 
training, we gathered about 500 positive samples, 1000 negative samples, and we 
used cascade classifier with 20 stages. On a standard PC, training took about 
\SI{12}{\hour}.

The best detection results are obtained at distances between \SI{1}{\metre} and 
\SI{5}{\metre} from the observer. After non-maximum suppression, a bounding box for 
each detection is computed and projected to egocentric world coordinates using 
the calculated extrinsic camera matrix.

\subsection{Heading Estimation}
\seclabel{heading_estimation}

All robots are visually identical, and we did not use the localization calculated by
the observed robots. The robot heading relative to the observer is used as primary cue for robot
identification. For visual heading estimation of each robot torso, which needs to be invariant to
leg orientations, we analyze the features of the upper half of the detected bounding boxes. We formulate 
this problem as a multiclass classification problem that was solved using an SVM multiclass classifier with
an RBF kernel. The full heading range was partitioned into ten classes of size \SI{36}{\degree}.

A dense HOG descriptor was applied on the grayscale channel and on the ``H'' 
channel in HSV color space. The resulting feature vector, plus the normalized center position of the bounding box are
forwarded to the SVM classifier. Note that we included the center position because visual features of the robot are different
depending on robot's position in the observer camera coordinates. For implementation, the LIBSVM 
library~\cite{CC01a} was used with k-fold cross-validation and grid search for tuning hyperparameters. In our 
experiments, the average error for heading estimation 
was \SI{17}{\degree}.

%%%%%%%%%%%%%%%%%%%%%%%%%%%%%%%%%%%%%%%%%%%%%%%%%%%%%%%%%%%%%%%%%%%%%%%%%%%%%%%%
\section{Tracking and Identification System}
% Story for how i divide the problem in data association and filtering
Tracking targets in the image plane is very popular and straightforward, but the 
often simple motion models break very easily and have an unpredictable effect 
when the camera or the observer moves. It is quite difficult to find a reliable motion model that works well in the different 
regions of the image. To address this issue, we propose tracking the target in 
egocentric world coordinates. By doing this, we can separate our problem into 
two different tasks. First, we need to identify each detection and then we can 
update the tracked positions of the robots based on the identification 
probabilities. Note that this is an entirely different setup than what we 
previously proposed~\cite{farazireal}, which was 
tracking in the image plane, followed by identification for the existing tracks.

% Story 
Data association is the most challenging component of the multi-target tracking 
problem. Although greedy solutions like the Hungarian method lead to an 
acceptable result with a low computational cost, they do not work well in 
challenging situations, especially in the case of occlusions. JPDA-like 
algorithms, which jointly consider all possible assignment hypotheses, and form 
an NP-hard problem, are too computationally expensive to be used in real-time 
applications. Hence, we need to use a suitable approximation to obtain both the 
required accuracy and efficiency. RNNs, in particular LSTM networks, are very 
powerful in capturing spatial and temporal dependencies in input/output data 
sequences. These characteristics are achieved by using non-linear 
transformations and hidden-state memory built into the LSTM cells.

% Why not prev works
We extend the method proposed by Milan et al.~\cite{Milan:2017:AAAI_RNNTracking} 
for data association. They suggested a two-layer LSTM network with 500 units for 
data association in the form of a single network that is used multiple times to 
process multiple detections. Although this architecture has the advantage of 
being able to cope with variable numbers of detections, simply by applying the 
network multiple times, it needs the predicted position of each target at each 
time. Note also that while the number of detections can vary, the number of 
targets must still be pre-selected with this architecture. Another requirement 
of their architecture is the need to manually choose a metric, in their case the 
Euclidean distance, and compute a pair-wise distance matrix $C \in 
\mathbb{R}^{M\times N}$ between the measurement and the predicted state of the 
target. A downside of this approach is that each detection is associated 
independently, so some potentially valuable information is discarded. In this 
paper we propose a new end-to-end architecture for data association that 
addresses the aforementioned issues.

\subsection{Proposed Architecture}

%-----------------------------------
\begin{figure}[!tb]
\vspace*{1ex}
\centering
\hspace*{-0.3cm}
\includegraphics[width=11.7cm]{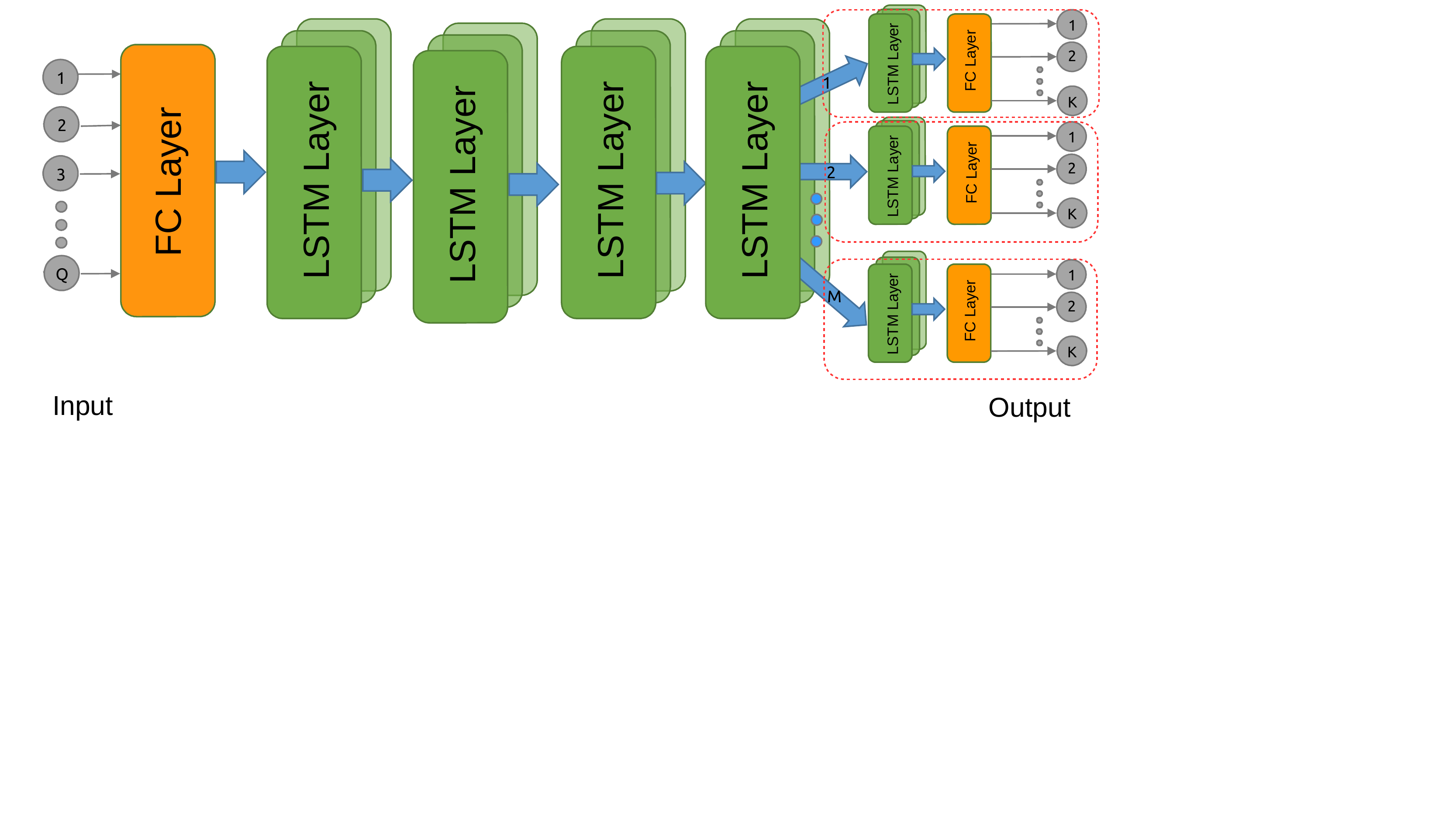}
\vspace*{-23ex}
\caption{The proposed deep network with five LSTM layers. Each output submodule is shown with a red dashed box.}
\figlabel{NN}
\vspace*{-1.8ex}
\end{figure}
%-----------------------------------

In the proposed architecture we seek to avoid imparting prior information into 
the system through the choice of a fixed motion model, like in the Kalman filter 
\cite{farazireal}. The use of a secondary network for state prediction~\cite{Milan:2017:AAAI_RNNTracking} is also
avoided to keep the system unified, and to only require a single loss function
and training set. Unlike Haarnoja et al.~\cite{haarnoja2016backprop}, who proposed
an approach for single object tracking, we gave the network the ability to define 
the content of its hidden states without using a hand-crafted network architecture.
Another consideration while designing the network was the choice of a deep network to give it the 
opportunity to learn hierarchical representations. 

We formulated the data association problem as a classification task that is 
performed on the maximum possible number of targets. We used a single network 
architecture, instead of multiple LSTM networks for different targets, because 
we wanted to leverage the codependency of the target assignments. The five layer 
proposed LSTM network is depicted in~\figref{NN}. Each layer contains 64 LSTM 
units. Note that by using one sub network for each target, the assignment 
probability matrix would be calculated independent from other target selections.

In our problem, we define the input vector $I_t \in  \mathbb{R}^Q$, where $Q = N + 
M(D+1)$, as the vector containing all available and observable states of the 
robots, where $N$ is the number of robots broadcasting their absolute 
headings, and $M$ is the maximum number of detections at each instant in time 
$t$, with the assumption that $M \geq N$. Observable states for each detection 
are the position ($x$,$y$) in normalized egocentric world coordinates, and the 
normalized absolute estimated orientation ($\phi$), such that 
$D=|\{x,y,\phi\}|=3$. In addition to $D$, we input the detection probability 
$\gamma \in [0,1]$ generated by the visual detector, which is necessary to be 
able to deal with the unknown number of detections and robots that are present. For training phase, we do put the detections in random order, and for inference phase, the detections can have random order as well. Note 
that if for any reason the detector outputs more than $M$ detections, the $M$ 
most probable detections are fed into the network. Extending $D$ to incorporate 
other perceived variables such as velocity or robot appearance,--- in case the 
robots were different,--- would be straightforward. In summary, the input to the 
network contains all the detections and the corresponding confidences and 
broadcasted heading from each of the robots.

The $y$ component of the robot locations, which corresponds to the rows of the 
image, is more accurately estimated than the $x$ component for two main reasons. First, the estimation of $x$ is more 
sensitive to errors in the projection from image coordinates to egocentric world 
coordinates. Second, detected bounding boxes in general were observed to produce 
more error in the $x$ direction than in the $y$. This justifies the claim that 
inputting pure position and orientation to the network is a better choice than 
calculating the Euclidean distance manually, because the network then has the 
chance to learn the correlations between the dimensions and effectively use 
something similar to, for example, the Mahalanobis distance. Another reason 
against the use of the manual Euclidean distance is the periodic nature of the 
$\phi$ component.

Each detection can either correspond to one of the robots or a false positive. This forms a 
total of $K$ different situations, where $K= N+1$. We used these $K$ possible valid outputs as different classes for each of the detections in an output submodule. Each class in the output submodules is encoding one possible association for each detection being assigned to
one of the $N$ robots, or being a false positive. The network learns to limit the assignments, such that each of the robots can be assigned to at most one detection.

As the loss function for each of the output submodules, we used the common negative log-likelihood of true scene 
state given the input. That is,
\begin{equation}
\hat W= \mathrm{argmin}_W -\sum_{i=1}^{\rho} \log P(O_i|I_{i};W) + \lambda\sum_{j=0}^{d}W_j^2
\end{equation}
where $O_i$ is the desired output for input $I_i$, $W$ is the weight matrix, $d$ is the length of $W$, and 
$\lambda$ is the regularization coefficient.

Among all different variations of LSTMs, we used the one that was used in~\cite{zaremba2014learning}.
The LSTM update formula for time step $t$ is,
\begin{equation}
\begin{split}
i_t &= \sigma(W_{xi}x_t + W_{hi}h_{t-1} + b_i)\\
f_t &= \sigma(W_{xf}x_t + W_{hf}h_{t-1} + b_f)\\
o_t &= \sigma(W_{wo}x_t + W_{ho}h_{t-1} + b_o)\\
g_t &= \tanh(W_{xc}x_t + W_{hx}h_{t-1} + b_c)\\
c_t &= f_t \odot c_{t-1} + i_t \odot g_t\\
h_t &= o_t \odot \tanh(c_t)
\end{split}
\end{equation}
where  $x_t$ is the input, $h_t$ is the hidden unit, $f_t$ is the forget gate, $i_t$ is the input 
gate, $o_t$ is the output gate, $c_t$ is the memory cell, and $g_t$ is the input 
modulation gate. Note that $\sigma(x)$ and $\tanh(x)$ are sigmoid and hyperbolic 
tangent nonlinearity for squashing gates to the respective range. As it depicted 
in~\figref{LSTM}, $x_t$, $h_{t-1}$, and $c_{t-1}$ are inputs to each LSTM cell.
%-----------------------------------
\begin{figure}[!tb]
\vspace*{1ex}
\centering
\hspace*{1.55cm}
\includegraphics[width=13cm]{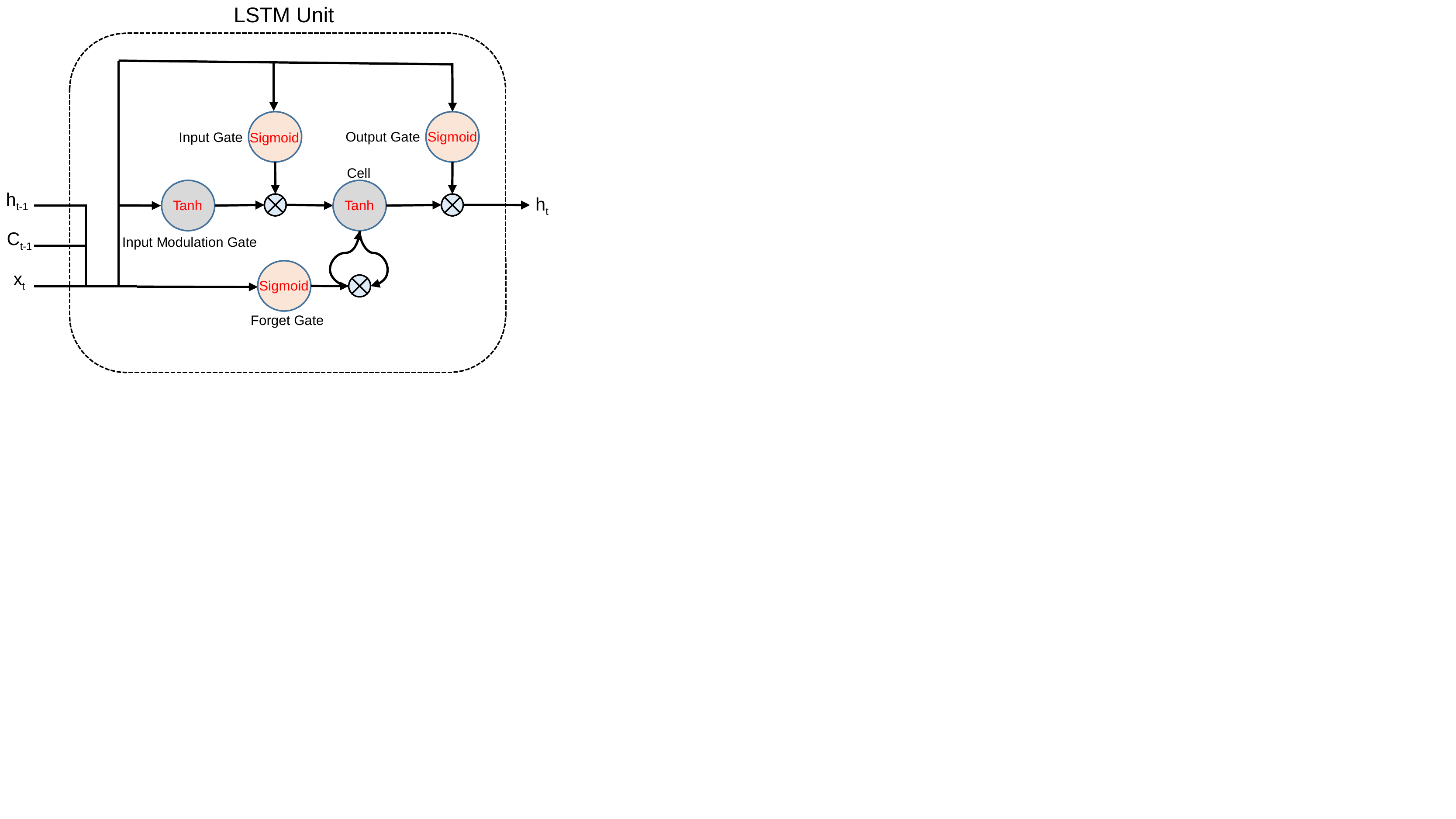}
\vspace*{-27ex}
\caption{The LSTM memory unit used in this paper.}
\figlabel{LSTM}
\vspace*{-2.9ex}
\end{figure}
%-----------------------------------
\subsection{Training Data}
Deep networks need a large amount of training data in order to converge to a 
solution without overfitting and while still generalizing well to previously 
unseen samples. Gathering a huge dataset like ImageNet~\cite{ILSVRC15} with 
ground truth labels, is very time-consuming and next to impossible in our setup. 
One solution that works in an unsupervised fashion for multi-tracking using deep 
learning, if raw input coordinates are used, has been proposed by Ondruska et al.~\cite{ondruska2016deep}, but in our situation that solution is not 
applicable because we need a discrete data association for identification. Even 
in the well-known problem of pedestrian tracking, there are a very limited 
number of datasets. Another solution might be the synthetic generation of data 
by sampling distributions generated from real data, as proposed by Milan et al.~\cite{Milan:2017:AAAI_RNNTracking}. They used two features, start position, and 
average velocity. Although this is a good idea to extend pedestrian tracking 
datasets, it still needs a considerable amount of samples to model different 
realistic motions.

We used a 2D simulator that can realistically simulate our problem and generate 
an unlimited number of sequences. In the simulator, we can specify different 
velocities and accelerations for the position ($x,y$) and rotation ($\phi$). We 
did not restrict our motion model, so effectively the robots were assumed to 
walk omnidirectionally. To simulate perception noise, we added Gaussian noise to 
resemble detection and projection noises. Note that to simulate more realistic 
data samples, we added more perception noise in the $x$ direction than in the 
$y$. $\phi$ estimation noise was calculated and utilized with similar statistics 
to our visual estimation algorithm \secref{heading_estimation}. Occlusions and 
walking out of the field of view is simulated as well. Moreover, false positives 
and false negatives similar to the detector's characteristics are simulated. 
Two screenshots of our simulator are shown in~\figref{SCREENSHOT}. 
To force the network to learn the actual relations between unordered detection inputs, we 
randomly indexed the detections in both training and inference phase. It was important to reset the 
cell states and hidden states after each simulated training sequence, to prevent 
learning some relations which were not intended due to backpropagation through 
time.

%-----------------------------------
\begin{figure}[!tb]
\vspace*{1ex}
\centering
\hspace*{0cm}
\includegraphics[width=8.5cm]{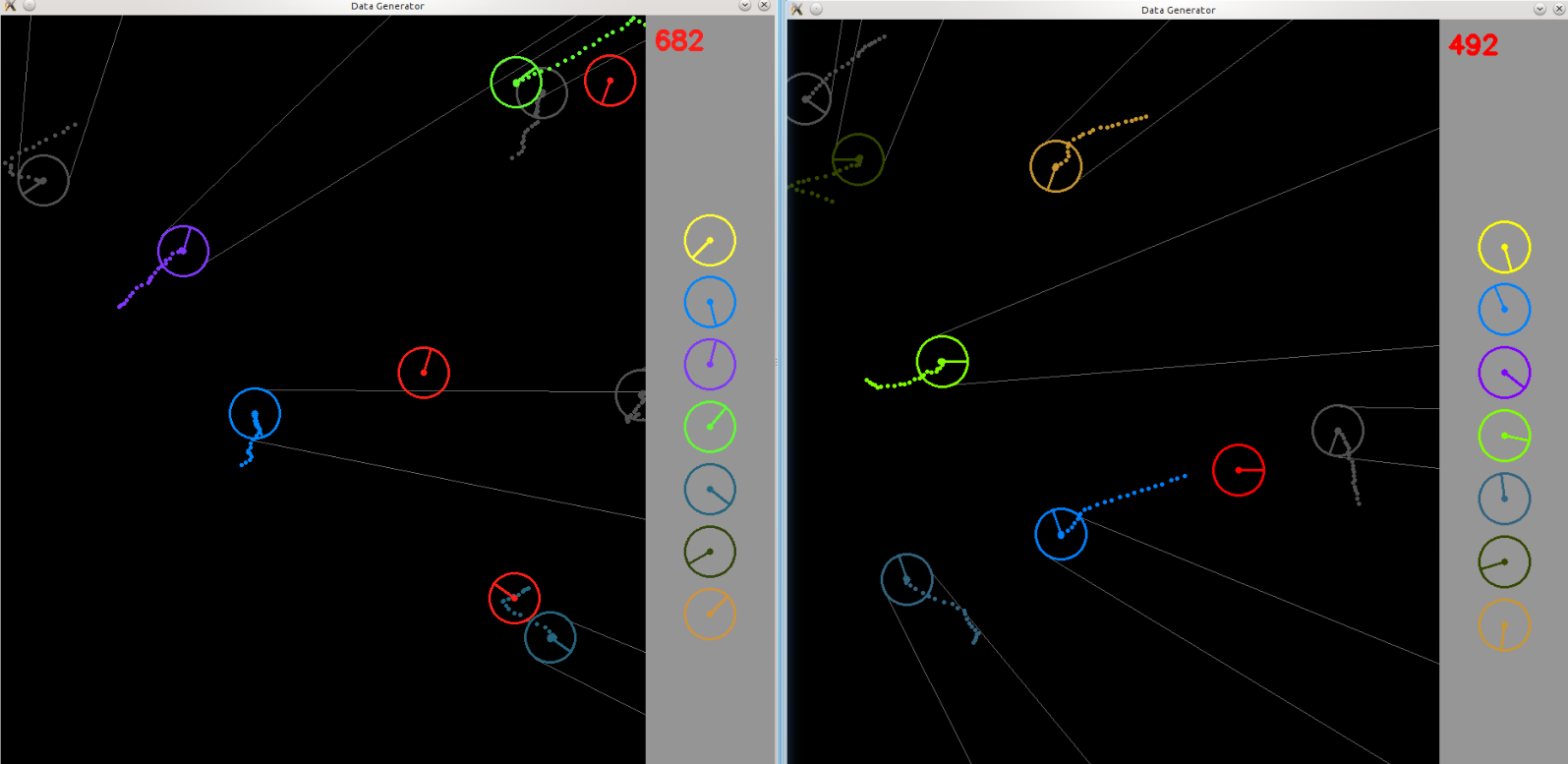}
\vspace*{-3ex}
\caption{Two screenshots of our simulator for $M=10$ and $N=7$.}
\figlabel{SCREENSHOT}
\vspace*{-3ex}
\end{figure}
%-----------------------------------

\subsection{Hyperparameters}
There is still no known proper solution for selecting correct hyperparameters 
for LSTMs~\cite{greff2016lstm}. To find a suitable set of hyperparameters, we 
did cross-validation and random search over log space~\cite{bergstra2012random}. 
These parameters were network size, depth, and learning rate. We used the Adam 
optimization method~\cite{kingma2014adam} for training. The learning rate started with 0.004 
and decayed with a rate of 0.0001. To regularize the network, we used $L_2$ 
regularization. Note that for training of the network, we cannot shuffle the sequences of the 
dataset because the network is learning the sequential relations between inputs 
and outputs. We used $\rho = 150$ previous steps in the memory for 
backpropagation through time. This is approximately 5 seconds at 
\SI{30}{\hertz}. In our experience, adding dropout reduces the performance of the network.

The training dataset was divided into mini-batches of sequences of size 500. We used the 
popular zero-mean and unit variance input data normalization. Training was 
performed on a computer which was equipped with four Titan X GPUs and a 32-core 
CPU. Multi-GPU training took two days on the synthetic data. For getting the 
best performance in the real setup, we used one of our recorded real sequences 
for fine-tuning of the network. This process took only one hour and boosted our 
performance in real experiments.

\subsection{Filtered Locations}
For tracking the real position of the detected robots, we use a first-order low pass filter to smooth each of the robots 
position relative to the observer, using the probability that comes from the 
last layer of the network. For updating each robot location we use the following formula.
\begin{equation}
T_i^j= \alpha L_i^k + (1-\alpha)T_{i-1}^j
\end{equation}

Where $\alpha$ is the smoothing factor for the measurement update, which is the likelihood of the classification for each of the robot.
$T_i^j$ is the location of the robot $j$ in current frame $i$ and $L_i^k$ is the location of detection $k$ which the network assigned to target $j$.
 By doing this, we can track the position of the 
robot in egocentric world coordinates of the observer very robustly. The 
calculated location is then broadcasted to each robot for further use.

%%%%%%%%%%%%%%%%%%%%%%%%%%%%%%%%%%%%%%%%%%%%%%%%%%%%%%%%%%%%%%%%%%%%%%%%%%%%%%%%
\section{Experimental Results}

\begin{figure}[!t]
\vspace*{0ex}
\centering
\begin{picture}(120,50)% width and height of the picture
\put(0,0){\includegraphics[width=4.2cm]{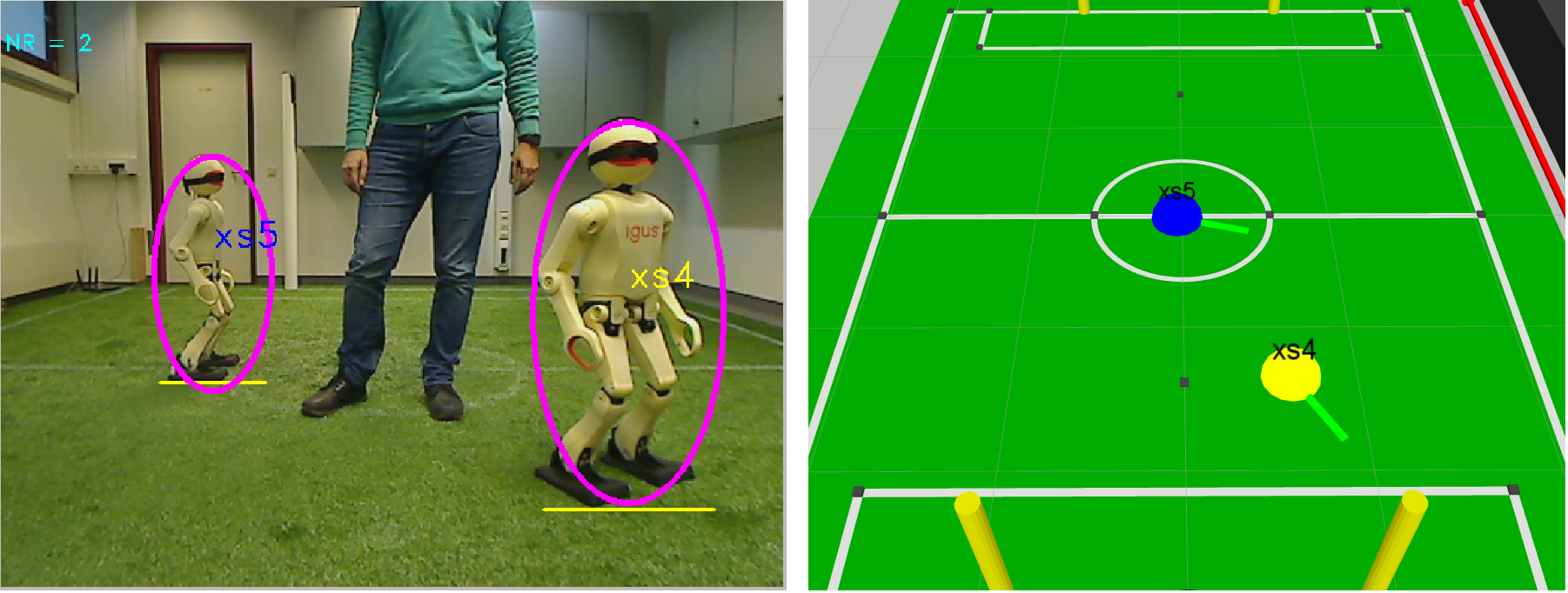}}
\put(2,35){frame 568}
\end{picture}
\begin{picture}(120,50)% width and height of the picture
\put(0,0){\includegraphics[width=4.2cm]{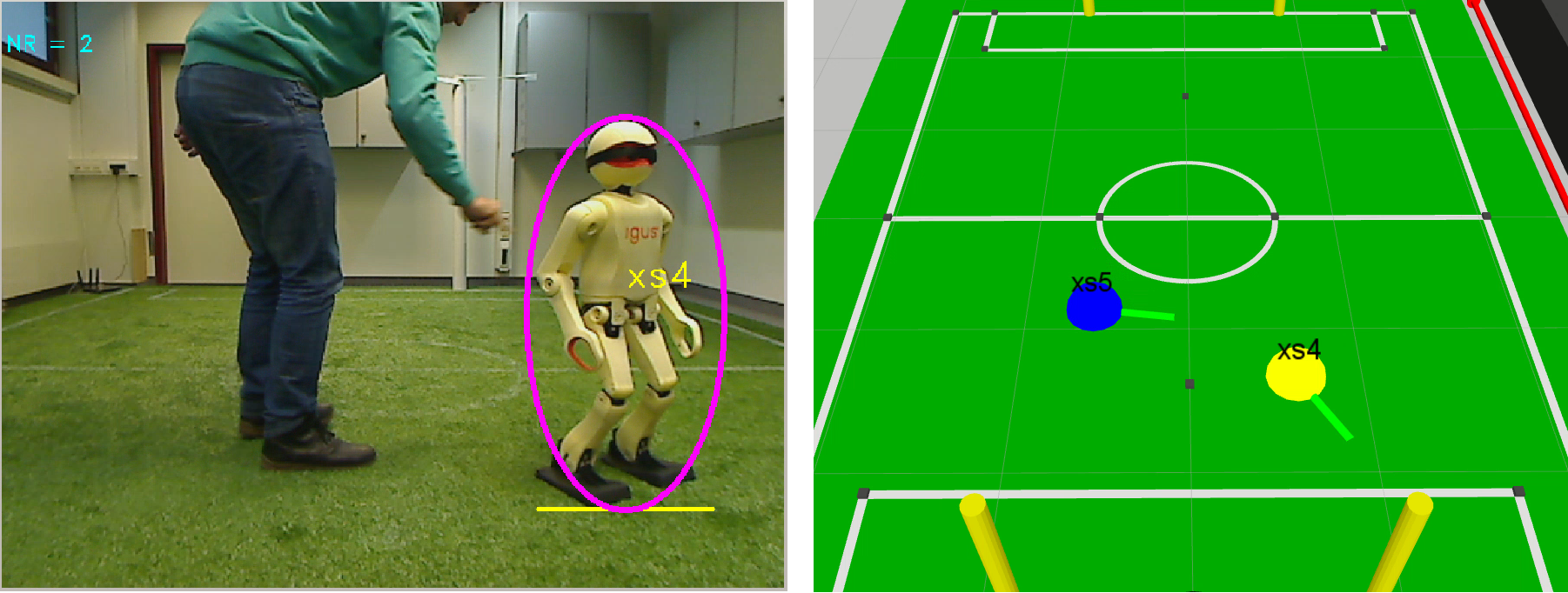}}
\put(2,35){frame 623}
\end{picture}
\begin{picture}(120,50)% width and height of the picture
\put(0,0){\includegraphics[width=4.2cm]{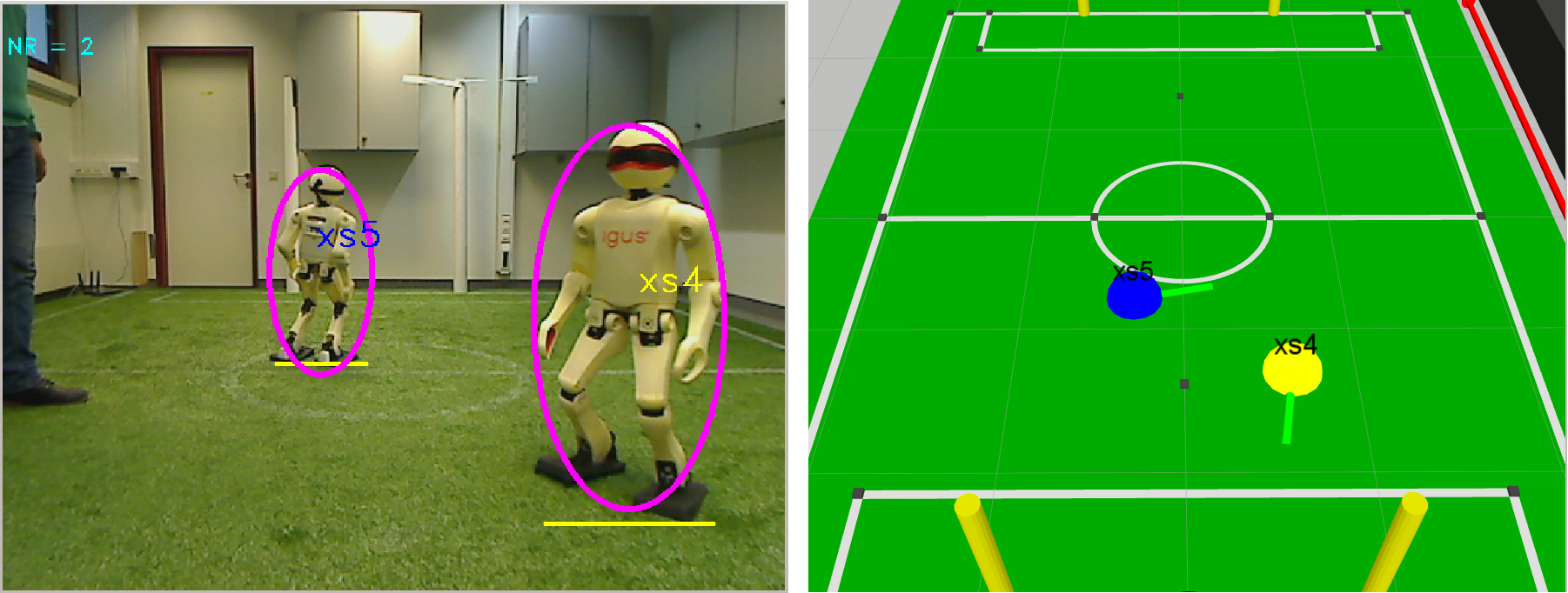}}
\put(2,35){frame 740}
\end{picture}
\begin{picture}(120,50)% width and height of the picture
\put(0,0){\includegraphics[width=4.2cm]{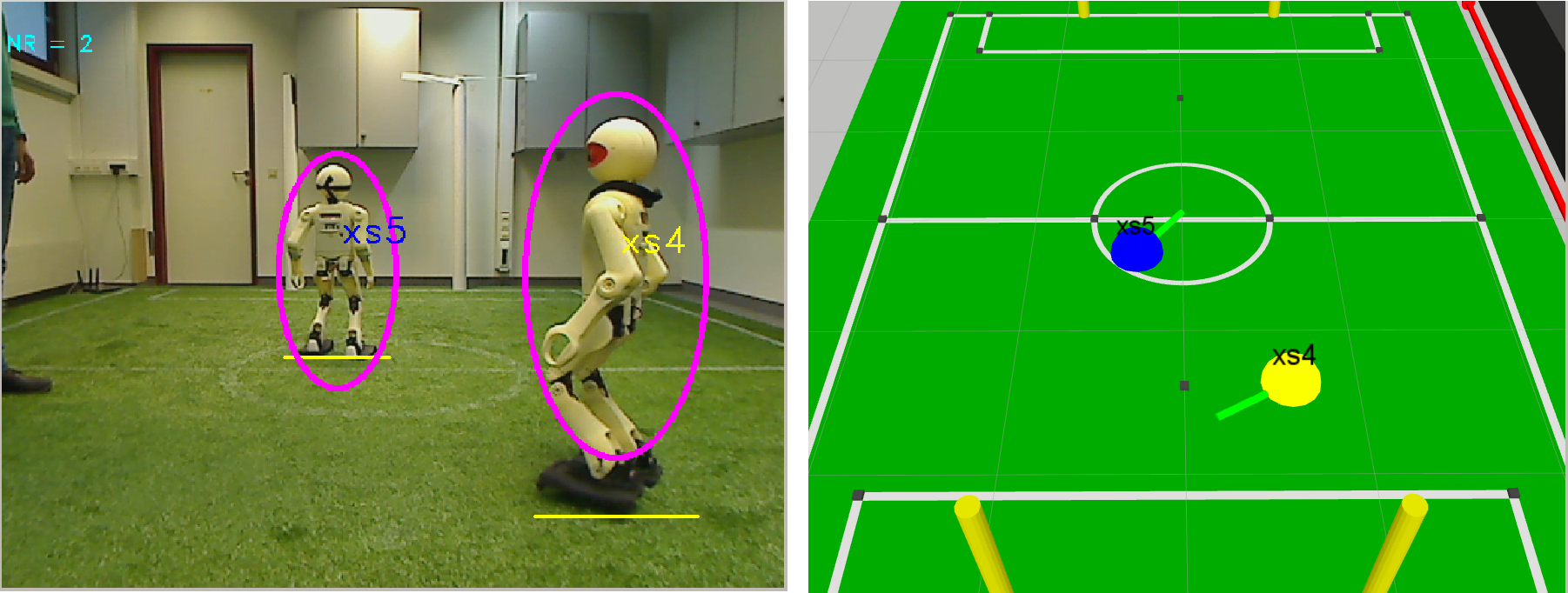}}
\put(2,35){frame 823}
\end{picture}
\begin{picture}(120,50)% width and height of the picture
\put(0,0){\includegraphics[width=4.2cm]{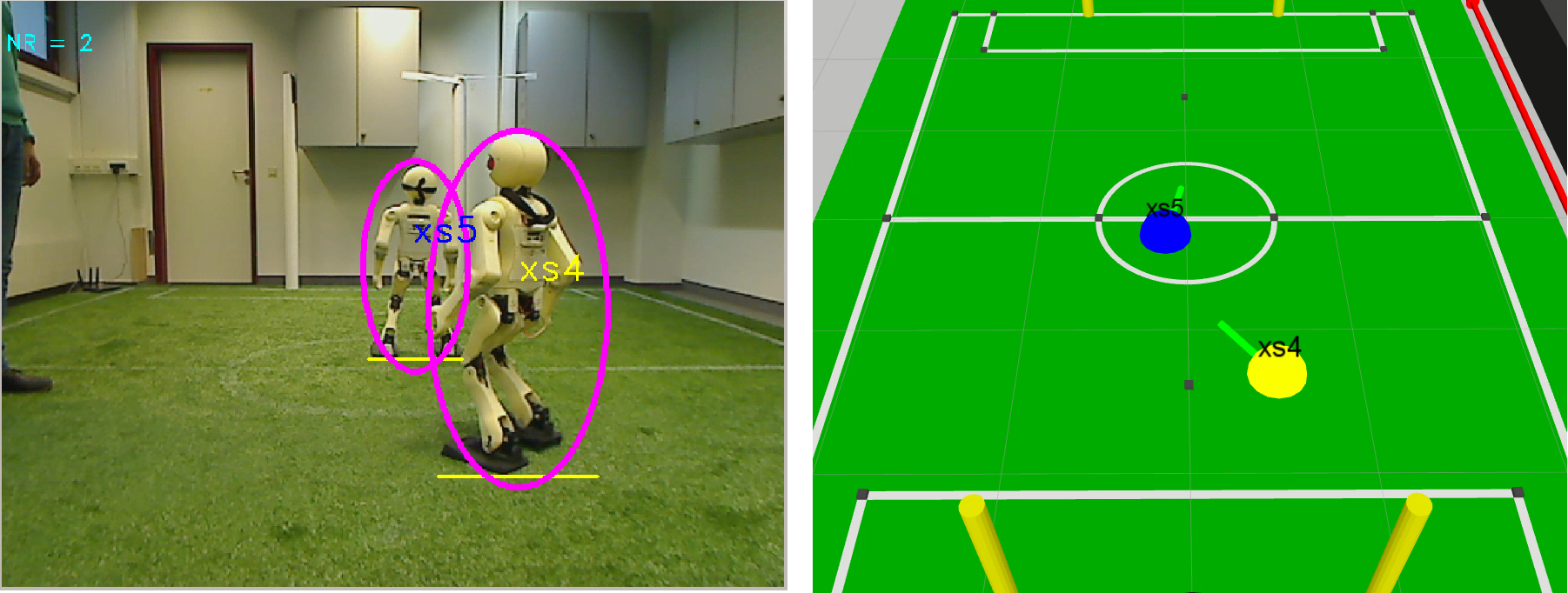}}
\put(2,35){frame 861}
\end{picture}
\begin{picture}(120,50)% width and height of the picture
\put(0,0){\includegraphics[width=4.2cm]{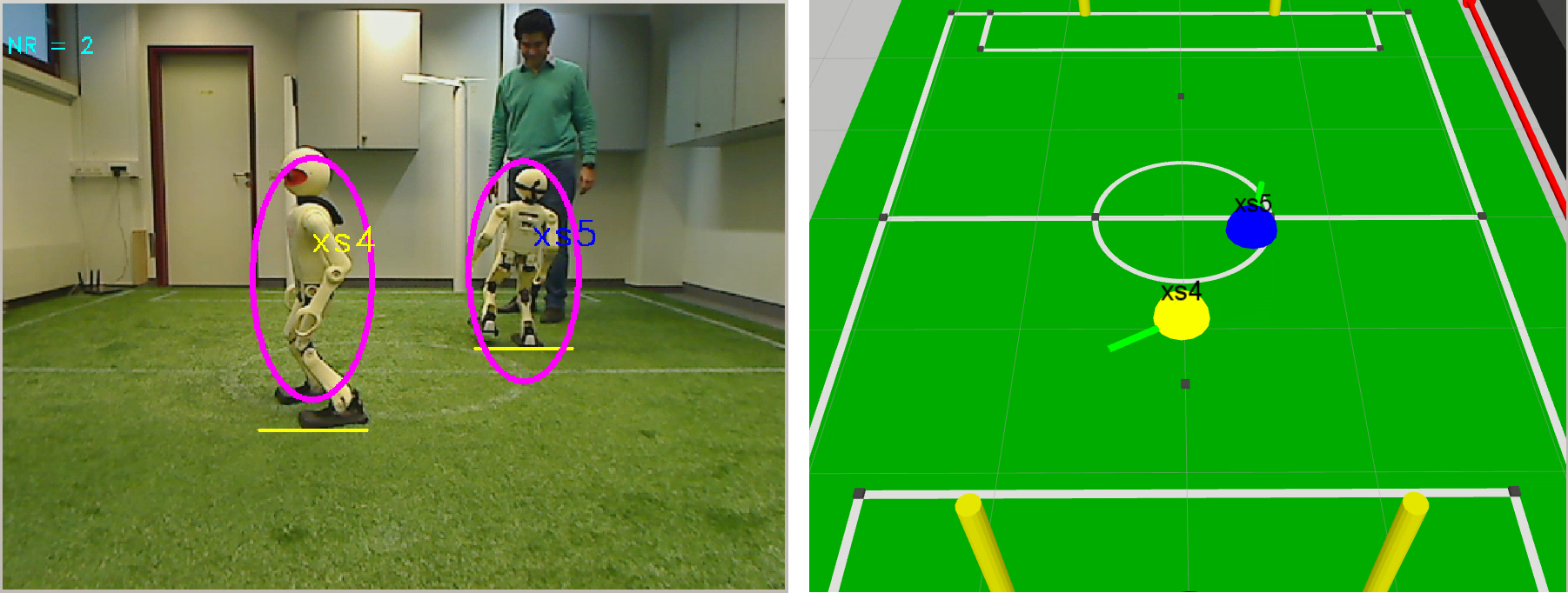}}
\put(2,35){frame 1012}
\end{picture}
\begin{picture}(120,50)% width and height of the picture
\put(0,0){\includegraphics[width=4.2cm]{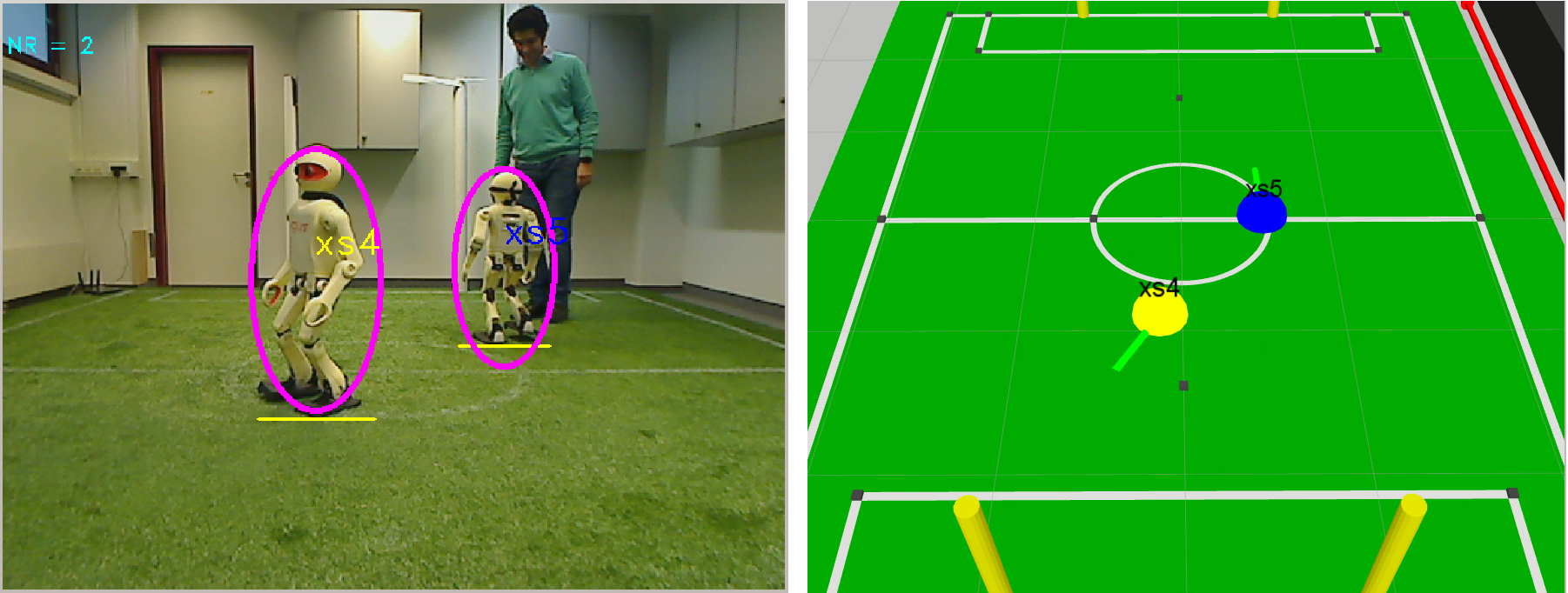}}
\put(2,35){frame 1077}
\end{picture}
\begin{picture}(120,50)% width and height of the picture
\put(0,0){\includegraphics[width=4.2cm]{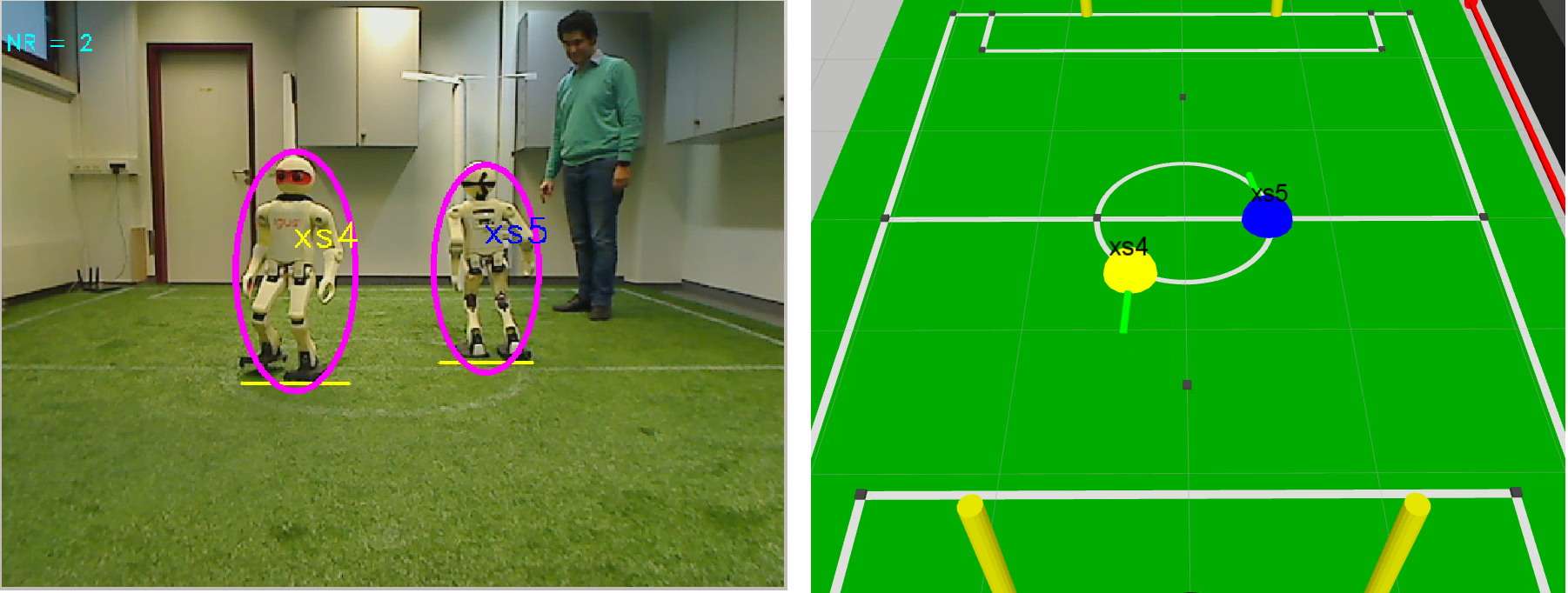}}
\put(2,35){frame 1189}
\end{picture}
\vspace*{-2.5ex}
\caption{Detection, tracking, and identification results obtained by our system on a single sequence.}
\figlabel{result}
\vspace*{-3.5ex}
\end{figure}
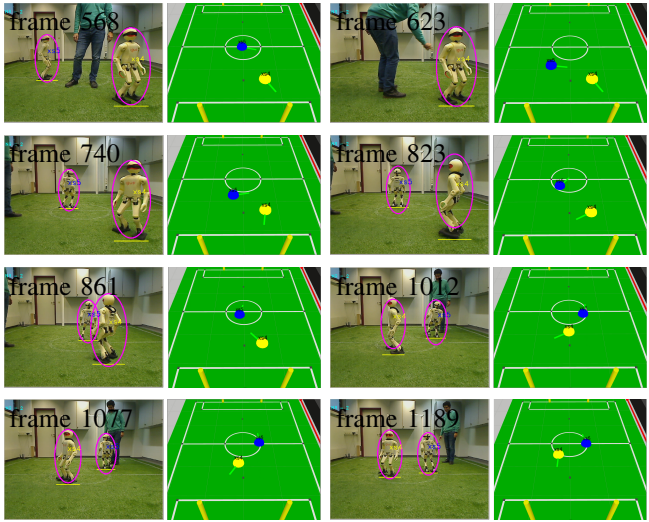
% warmup

Due to the unique setup of our problem, there is not publicly available 
benchmark that we can demonstrate our method on. We instead compared our approach
with three commonly used baseline methods, on our own collected datasets. 
For each of these baselines, we first detect the robots and form 
tracks as described by ~\cite{farazireal}. Then we associate these tracks to the robots and perform robot 
identification. 
In all these baselines, we applied the widely used Kalman filter with a constant 
acceleration model. The main difference between them is their data association method and techniques
for maintaining the tracks. Similar to Milan et al.~\cite{Milan:2017:AAAI_RNNTracking}, the bipartite matching for the Kalman-HA 
method is solved using the Hungarian algorithm without any heuristics, time 
delay or post-processing. Tracks are initiated and terminated as soon as the detection appears or is missed.
The Kalman-HA2 method extends this with a set of 
heuristics to handle false positives and false negatives in an additional 
post-processing step, similar to~\cite{farazireal}. The JPDA method used as the 
third baseline is inspired by the JPDAR method from~\cite{hamid2015joint}, but 
without using the m-best approximation in order to get the full capacity of 
JPDA. The tracks are maintained using the JPDA algorithm, and the linear 
assignment problem, which is a necessary step for identification, is solved 
using the Hungarian algorithm. In all experiments, the success rate is calculated by counting the total number of correct assignment and dividing it by the total number of detections for the entire sequence.
\begin{table}[!t]
\caption{Robot identification results on the synthetic dataset.}
\renewcommand{\arraystretch}{1.1}
\centering
\begin{tabular}{| l | c c c|}
\hline
Setup & \:$M$=3, $N$=2\: & \multicolumn{1}{|c|}{\:$M$=5, $N$=3\:} & \multicolumn{1}{|c|}{\:$M$=10, $N$=7\:} \\
\hline
Human & 94.3\% & 86.3\% & \textbf{67.3}\% \\
\hline
Kalman-HA & 75.6\% & 72.2\% & 53.1\% \\
\hline
\textbf{ours} & \textbf{96.2}\% & \textbf{87.1}\% & 66.5\% \\
\hline
\end{tabular}
\tablabel{simres}
\end{table}

\subsection{Simulation Experiment}
To demonstrate the capability of our proposed approach, we first performed 
experiments on simulated data using three different setups---two targets with up 
to three detections, three targets with up to five detections, and seven targets 
with up to ten detections. The dataset used for testing was different from 
datasets for training and validation. We tested Kalman-HA on our simulated 
dataset, and measured over 10 different sequences of 1000 frames each. These 
results are shown in ~\tabref{simres}. We also chose four of those sequences 
randomly and ask four different persons to solve the association problem by 
clicking on the right color choice for each detection.For human-level performance experiment, we asked 
the participant to select the correct associations between the detections and 
targets in each frame. The user can see the reported heading from the robot as 
well as the detected location and orientation. A correct association selection 
is done when the color of each detection matches the corresponding robot color. 
By left/right clicking on each detection, the user can change the color in 
forward or reverse order. We observe a superior result of our method compared to 
human-level performance on the three and five detection setups, and also a 
comparable result on the ten detections setup. Also observe superior results on 
all cases of our method in comparison to Kalman-HA. Note that the other two 
baseline methods cannot easily be tested on the simulated dataset because they 
are purely formulated in image coordinates, as opposed to the required 
normalized egocentric world coordinates. In the simulated dataset, we observed 
that the typical case of failure was at frames in which randomly generated 
detections were spatially close to one of the targets. Our results on the 
simulated dataset for wide range of target number emphasizes the scalability of 
our network.

\subsection{Real-Robot Experiment}

Including the observer, three \iguhop robots~\cite{allgeuer2015child} were used 
for the verification of the approach on real-world data. In our setting, the observer
was the goalkeeper observing the reset of the team. For performing real-world tests,
we took the model that was trained on simulated data, and fine-tuned
it on real-world data captured by the observer robot. The dataset used 
for testing was different from datasets for fine-tuning. Note that for a 
valid comparison in the real-world dataset, we used the same inputs coming from 
the robot detector for all methods. 

For each of three different sequence lengths, we tested the methods on four randomly chosen sequences
of that length. The chosen sequence lengths were 200, 
400 and 800, frames respectively. The methods were also evaluated on the entire dataset, for a total of 3140
frames. The collected dataset included partial, short term and long term occlusions, as well as varying lighting conditions. Our 
results on a frame sequence is shown in~\figref{result}. We observed a superior result of our method compared to all tested baselines. In ~\tabref{results} we 
report the average success rates for the various methods and test cases. 

Although the average distance from the actual position of the robot is heavily 
dependent on the sensitivity of the projection, it is fair to compare results 
from different algorithms, if all of them share the same projection operation. Average distance is calculated by averaging all present robot location errors compared to the ground truth.
Our method gained less average localization error in comparison to the other 
methods as reported in ~\tabref{avgerror}.

\begin{table}[!t]
\caption{Average localization error for tracking.}
\renewcommand{\arraystretch}{1.1}
\centering
\begin{tabular}{| l | c c c c |}
\hline
Baseline & \:Kalman-HA\: & \multicolumn{1}{|c|}{\:Kalman-HA2\:} & \multicolumn{1}{c|}{\:JPDA\:} & \: \textbf{Ours}\\
\hline
Average error & \SI{0.67}{\metre} & \SI{0.30}{\metre} & \SI{0.29}{\metre} & \textbf{0.22\,m} \\
\hline
\end{tabular}
\tablabel{avgerror}
\end{table}

%ID switch

In a control experiment, we tested feeding sequences to the network in a random order 
and the results were quite poor, indicating that the network is using temporal 
relations between the inputs, and does not act like a feed-forward network.

A single forward pass of our network on the \iguhop took about \SI{4}{\milli\second} ($\approx$\SI{250}{\hertz}).
As another test, we replaced 
the LSTM cells with the more recent Gated Recurrent Units (GRUs) 
\cite{cho2014learning}. We did not observe any considerable benefit for our 
application.

%final notes
The fact that our proposed network can learn to solve the highly complex problem 
of data association based solely on learning is promising. We observed that when 
a target enters the field of view, instead of an instant increase in the 
total number of associated robots in the output, which is the case in greedy 
approaches like the Hungarian algorithm, the output set 
changes only after a few frames. This is a very useful feature to address false 
positives. Overall, this indicates that the recurrent 
architecture of the network can handle augmenting the set of 
visible robots well. Another crucial part of the system is the ability to 
recover from an incorrect data association. We tested the robustness of our 
network by forcing it to make incorrect data associations. In order to do so, we 
artificially swapped the input headings, and only restored them once the 
identifications had become incorrect. We observed that the network was reliably 
able to recover the correct solution. Note that if we ignore more 
difficulties for visual detection and visual heading estimation,  
we can use this method for moving observer if we know the motion model of the 
walking observer. This can be done by adding the observer motions to the 
detection locations. The network is trained on simulated random motions with 
random velocity and acceleration, so it can generalize to any unseen motion behavior. 
As we normalize the field dimensions in a preprocessing step, we can use the 
network on any field dimension. The network has the ability to learn long-term dependencies as well as
statistics of the detections and the relations between them. These are the main
reasons that our network outperforms other model-driven methods.

\begin{table}[!t]
\caption{Robot identification results on the real dataset.}
\renewcommand{\arraystretch}{1.1}
\centering
\begin{tabular}{| l | c c c c |}
\hline
Frames & \:200\: & \multicolumn{1}{|c|}{\:400\:} & \multicolumn{1}{c|}{\:800\:} & \:Total\:\\
\hline
Kalman-HA & 73.2\% & 75.5\% & 72.1\% & 73.8\%\\
Kalman-HA2 & 87.2\% & 84.0\% & 86.3\% & 85.5\%\\
JPDA & 87.1\% & 84.6\% & 85.6\% & 86.3\%\\
\hline
\textbf{Deep LSTM (ours)} & \textbf{89.8}\% & \textbf{90.3}\% & \textbf{92.4}\% & \textbf{91.1}\%\\
\hline
\end{tabular}
\tablabel{results}
\end{table}

% Balance out the columns on the last page
%\addtolength{\textheight}{-46mm}

%%%%%%%%%%%%%%%%%%%%%%%%%%%%%%%%%%%%%%%%%%%%%%%%%%%%%%%%%%%%%%%%%%%%%%%%%%%%%%%%
\section{Conclusion}

In this work, we proposed a practical pipeline for real-time visual tracking and identification of robots with the same appearance. Experimental results indicates that the proposed method can work well on simulated and real data and can cope with difficulties like long-term occlusions, despite a lack of visual differences between the robots. We achieved this by formulating the problem as a deep learning task and exploiting the sequences in the models, in the form of an RNN. The proposed system utilized heading estimation and spatial information for robot identification. Our system has applications in real-world scenarios including robot collaboration tasks, monitoring a team of robots, and cooperative localization and mapping. 

%%%%%%%%%%%%%%%%%%%%%%%%%%%%%%%%%%%%%%%%%%%%%%%%%%%%%%%%%%%%%%%%%%%%%%%%%%%%%%%%
\section*{Acknowledgment}
\seclabel{acknowledgment}
This work was partially funded by grant BE 2556/10 of the German Research Foundation (DFG).
The authors would like to thank Philipp Allgeuer for help in editing the article and assisting in performing experimental tests.
The authors would also like to thank Johannes K{\"u}rsch, Niloofar Azizi, and Donya Rahmati for performing the human-level experiments.

\IEEEtriggeratref{36} % Adjust this to balance the last two columns
\IEEEtriggercmd{\newpage}
\bibliographystyle{IEEEtran}
\bibliography{IEEEabrv,document}

%\addtolength{\textheight}{-12cm}

%%%%%%%%%%%%%%%%%%%%%%%%%%%%%%%%%%%%%%%%%%%%%%%%%%%%%%%%%%%%%%%%%%%%%%%%%%%%%%%%
\end{document}